\documentclass{article}
\usepackage{amsmath,graphicx,mlspconf}

\usepackage[utf8]{inputenc} 
\usepackage[T1]{fontenc}    

\usepackage{booktabs}       
\usepackage{amsfonts}       
\usepackage{nicefrac}       
\usepackage{microtype}      

\usepackage{amsmath}
\usepackage[flushleft]{threeparttable}

\usepackage{algorithm}  
\usepackage{algpseudocode}  

\usepackage[bookmarks=false]{hyperref}
\usepackage{multirow}

\title{A synthetic dataset for deep learning}
\name{Xinjie Lan}
\address{Anonymous}
\address{University of Delaware\\
	Electrical and Computer Science Department\\
	Newark, Delaware, 19711, USA}
\begin{document}
%

\maketitle
\begin{abstract}
In this paper, we propose a novel method for generating a synthetic dataset obeying Gaussian distribution.
Compared to the commonly used benchmark datasets with unknown distribution, the synthetic dataset has an explicit distribution, i.e., Gaussian distribution.
Meanwhile, it has the same characteristics as the benchmark dataset MNIST.
As a result, we can easily apply Deep Neural Networks (DNNs) on the synthetic dataset.
This synthetic dataset provides a novel experimental tool to verify the proposed theories of deep learning.

\end{abstract}
\section{Introduction}
\label{sec:intro}

Deep learning is a subset of machine learning algorithms that construct the Deep Neural Networks (DNNs) to solve complex problems \cite{CNN-Nature}.
Although it has achieved great success in various fields, such as speech recognition \cite{speech-recognition} and image classification \cite{CNN}, the internal logic of deep learning is still not convincingly explained and DNNs have been regarded as "black boxes" \cite{DNN_blackbox}.

Based on an underlying premise that DNNs establish a complex probabilistic model \cite{posterior1, pearl, cnn-posterior, posterior4}, numerous theories, such as the representation learning \cite{representation-dl, Goodfellow-et-al-2016, DRMM}, the Information Bottleneck (IB) theory \cite{Soatto1, DNN-Bottleneck, IP-argue, DNN-information}, 
have been proposed  to explore the working mechanism of deep learning.
Though the proposed theories reveal some important properties of deep learning, such as hierarchy \cite{representation-dl, Goodfellow-et-al-2016} and sufficiency \cite{Soatto1, DNN-information},
a fundamental problem is that the proposed theories cannot be directly validated by empirical experiments due to the fact that the distributions of the benchmark datasets, e.g., MNIST, are unknown.
For example, hierarchy is an important property of DNNs, but we still cannot explicitly formulate the hierarchy property and directly validate it by empirical experiments.

\pagebreak

\begin{figure}[!t]
\centering
\includegraphics[scale=0.55]{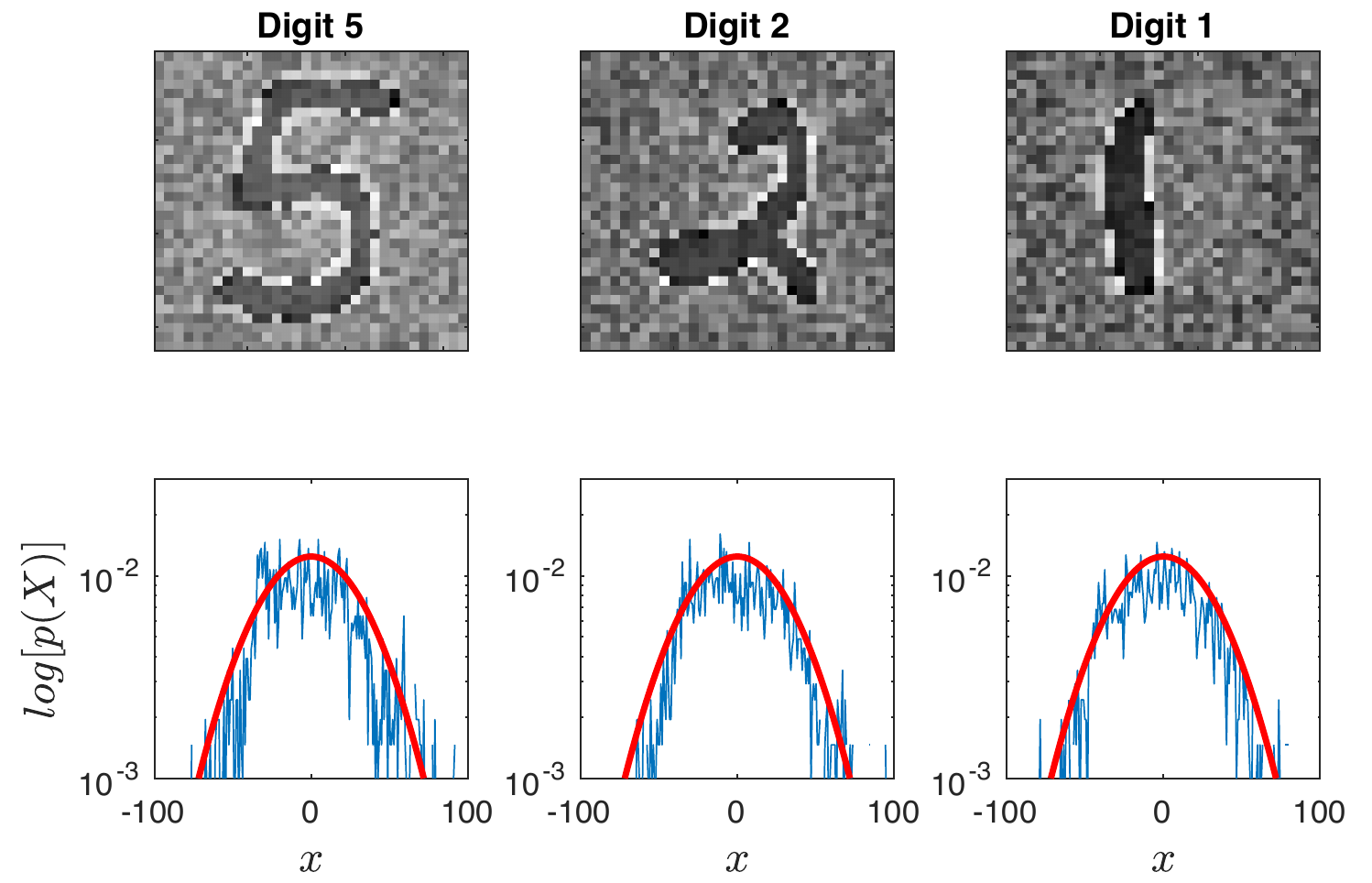}
\caption{\small{
The first row shows three synthetic images of handwritten digits, the second row shows their respective histograms, and the red curve indicates the Gaussian distribution $\mathcal{N}(0, 1024)$.
}}
\label{fig_synthetic_gaussian}
\end{figure}

To solve this problem, we propose a novel algorithm for generating a synthetic dataset obeying a Gaussian distribution based on the NIST \footnote{\url{https://www.nist.gov/srd/nist-special-database-19}} dataset of handwritten digits by class.
In particular, the synthetic dataset has the same characteristics as the benchmark dataset MNIST \cite{LeCun}.
Specifically, the synthetic dataset consists of 70,000 $32 \times 32$ grayscale images in 10 classes (digits from 0 to 9). 
Each class has 6,000 training images and 1,000 testing images. 
Fig. \ref{fig_synthetic_gaussian} shows three synthetic images.
Therefore, we can easily apply various DNNs on the synthetic dataset like MNIST.
Since all the grayscale images are sampled from a known distribution, the synthetic dataset obeys the Gaussian distribution.


This paper is organized as follows. Section 2 describes the specific method for generating the synthetic dataset obeying a known Gaussian distribution and Section 3 shows that the synthetic dataset can be easily applied to most commonly used DNNs. Section 4 demonstrates that given the synthetic dataset, we can verify some important properties of deep learning, e.g., hierarchy, based on the recent proposed probabilistic explanation of hidden layers of DNNs \cite{xinjie2, CRF_RNN}.

\pagebreak

\section{The method for generating the synthetic dataset}
\label{sec:format}

An underlying assumption of deep learning is that the given training dataset $\boldsymbol{\mathcal{D}} = \{(\boldsymbol{x}_n, \boldsymbol{y}_n)| \boldsymbol{x}_n \in \boldsymbol{R}^{S}, \boldsymbol{y}_n \in \boldsymbol{R}^{L}\}_{n=1}^{N}$ is composed of i.i.d. samples from a joint distribution $p_{\boldsymbol{\theta}}(\boldsymbol{X}, \boldsymbol{Y}) = p(\boldsymbol{Y|X})p(\boldsymbol{X})$, where $p(\boldsymbol{X})$ describes the prior knowledge of $\boldsymbol{X}$, $p(\boldsymbol{Y|X})$ describes the connection between $\boldsymbol{X}$ and $\boldsymbol{Y}$, and $\boldsymbol{\theta}$ indicate the parameters of $p_{\boldsymbol{\theta}}(\boldsymbol{X}, \boldsymbol{Y})$. 
Since we can easily formulate $p(\boldsymbol{Y|X})$ given $\boldsymbol{\mathcal{D}}$, $p(\boldsymbol{X})$ is the key of explicitly formulating $p_{\boldsymbol{\theta}}(\boldsymbol{X}, \boldsymbol{Y})$.

Unlike previous works using a complex probabilistic model to formulate $p(\boldsymbol{X})$ \cite{statistical_img, GSM-stat} for a given dataset, we first generate a random dataset obeying a Gaussian distribution and then use the generated random dataset to construct a synthetic image based on the mask derived from a benchmark dataset.
Since each data in the random dataset obeys $\mathcal{N}(0, 1024)$, we can conclude that the synthetic image also obeys $\mathcal{N}(0, 1024)$ based on the spatial stationary property, i.e., $\forall (i, j) \neq (m,n), p(\boldsymbol{X}(i, j)) = p(\boldsymbol{X}(m, n))$.

More specifically, the method includes seven steps:
 (i) generating a random vector $\boldsymbol{x} \in \boldsymbol{\mathbb{R}}^{1024\times 1}$ by sampling the Gaussian distribution $\mathcal{N}(0, 1024)$ for constructing a synthetic image with dimension $32 \times 32$;
 (ii) converting an image of the NIST dataset into a binary image;
 (iii) extracting the central part of the binary image and the dimension of the derived image is $64 \times 64$;
 (iv) downsampling the derived image in the previous step to obtain a binary image with dimension $32 \times 32$;
 (v) generating the mask of the binary digits image based on the Canny edge detection algorithm \cite{Canny}, and the mask indicates four parts of the binary image: outside, outside boundary, inside boundary and inside; 
 (vi) deriving an ordered vector $\boldsymbol{\hat{x}}$ by sorting $\boldsymbol{x}$ in the descending order and decomposing $\boldsymbol{\hat{x}}$ into four parts, i.e., $\boldsymbol{\hat{x}} = \{ \boldsymbol{\hat{x}}_1, \boldsymbol{\hat{x}}_2, \boldsymbol{\hat{x}}_3, \boldsymbol{\hat{x}}_4\}$, where $\boldsymbol{\hat{x}}_1$ corresponds to the outside, $\boldsymbol{\hat{x}}_2$ the inside boundary, $\boldsymbol{\hat{x}}_3$ the outside boundary, and $\boldsymbol{\hat{x}}_4$ the inside.
 (vii) generating a synthetic image by randomly placing each pixel in the four sub-vectors into a random position within the corresponding masks.

\begin{figure}[!b]
\centering
\includegraphics[scale=0.44]{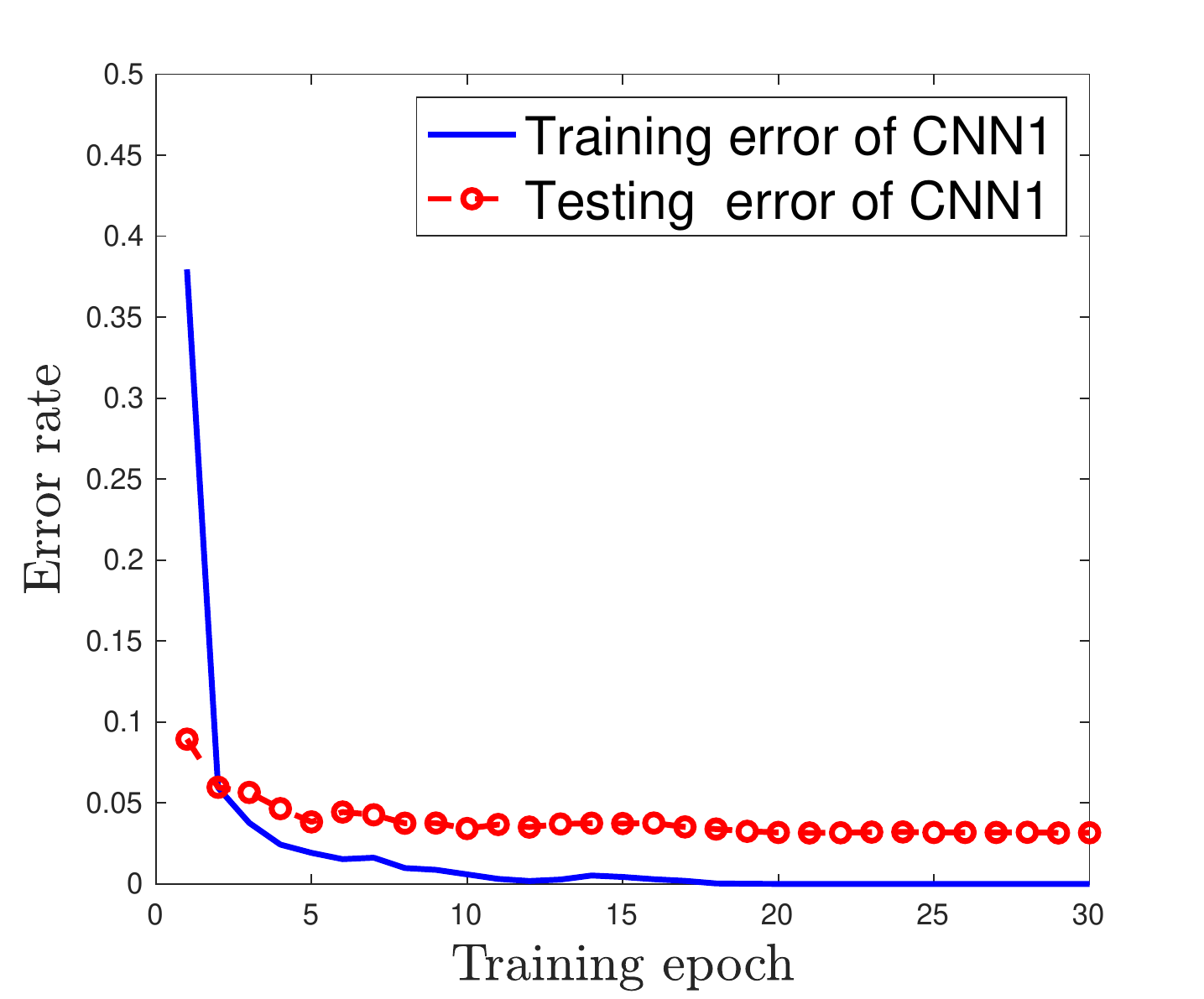}
\caption{\small{
The performance of CNN1 on the synthetic dataset
}}
\label{fig_cnn_synthetic}
\end{figure}

\pagebreak

\begin{figure*}[!t]
\centering
\begin{minipage}[b]{0.99\linewidth}
\centerline{\includegraphics[scale=0.7]{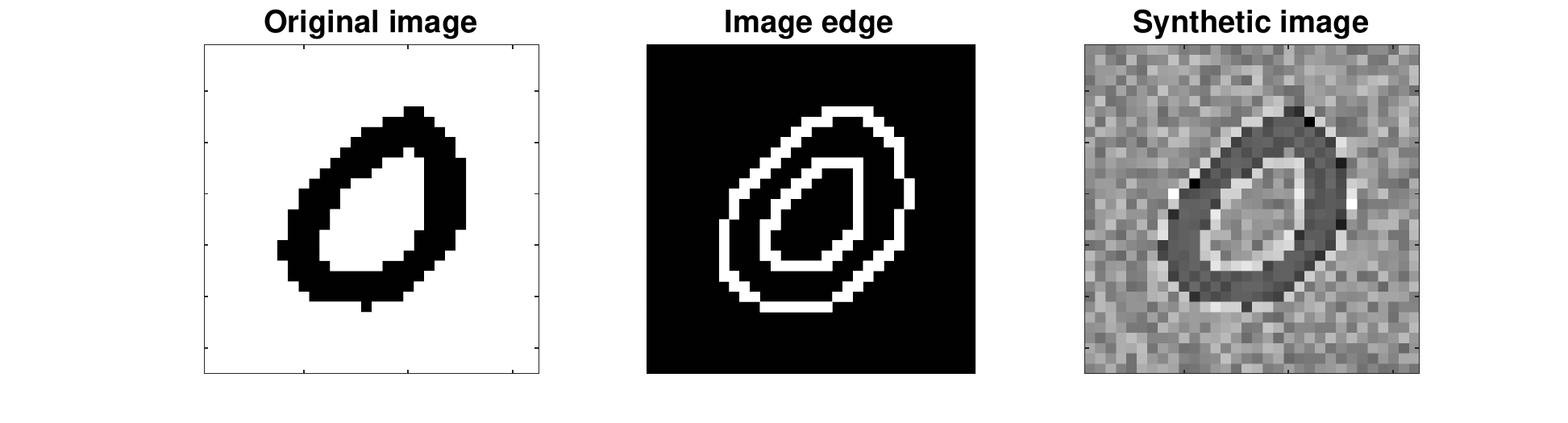}}
\end{minipage}
\vfill
\begin{minipage}[b]{0.99\linewidth}
\centerline{\includegraphics[scale=0.7]{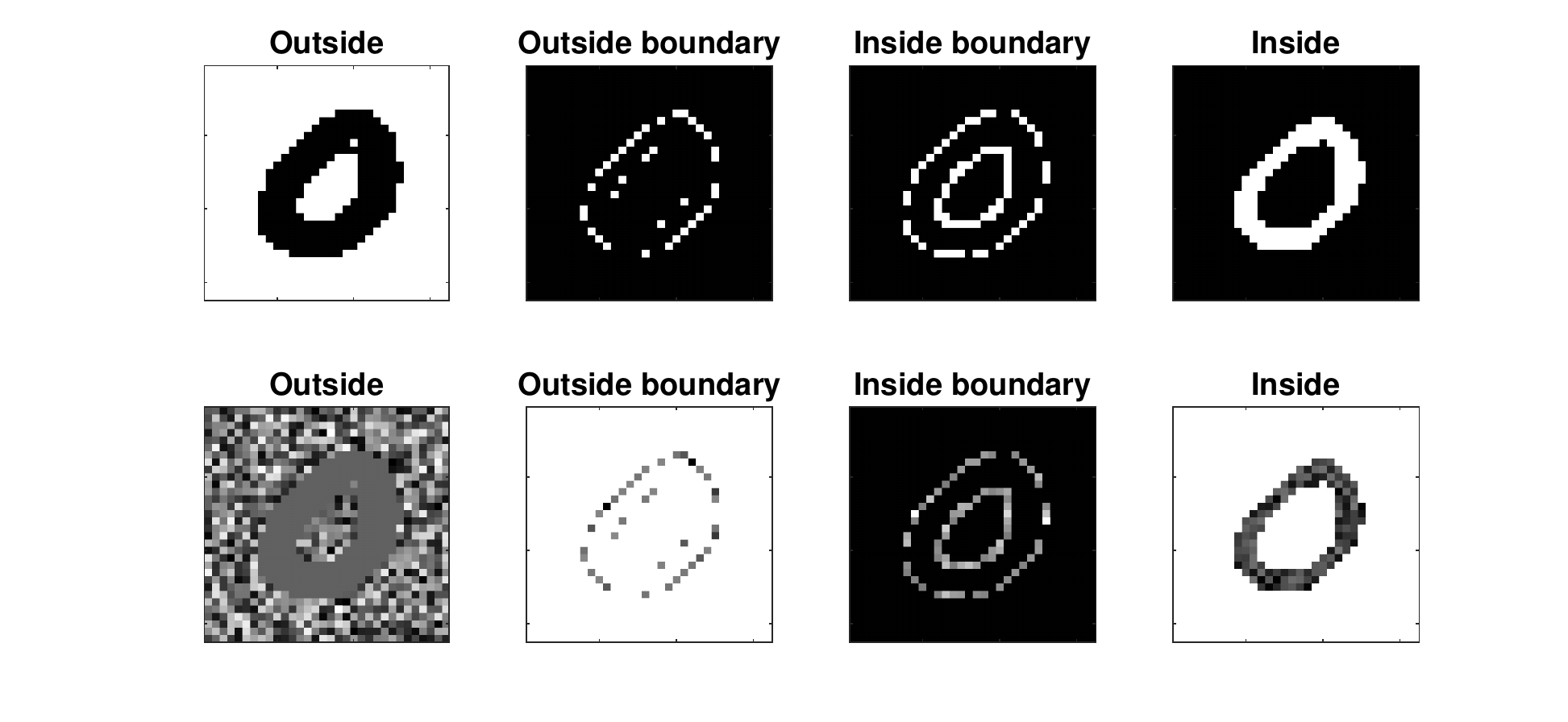}}
\end{minipage}
\caption{\small{ 
The first row shows an original image, its edge, and the corresponding synthetic image based on the original one. 
The second row uses white pixels to show the four parts of the mask image $\boldsymbol{m}_{cd}$.
The third row shows the synthetic image corresponding to each part of $\boldsymbol{m}_{cd}$.
}}	
\label{fig_synthetic_image}
\end{figure*}

The method for generating synthetic image is summarized in Algorithm 1, and Fig. \ref{fig_synthetic_image} visualizes the relationship between $\boldsymbol{\hat{x}} = \{ \boldsymbol{\hat{x}}_1, \boldsymbol{\hat{x}}_2, \boldsymbol{\hat{x}}_3, \boldsymbol{\hat{x}}_4\}$ and their corresponding masks.

 \begin{algorithm}[!t]  
  \caption{The algorithm for generating the synthetic dataset}  
   \label{bcnn_infer}  
  \begin{algorithmic}[1]  
    \Require
     NIST dataset of handwritten digits by class
    \begingroup
    \everymath{\footnotesize}
    \small
    
    \Repeat
      \State sampling $\mathcal{N}(0, 1024)$ to derive a random vector $\boldsymbol{x} \in \boldsymbol{\mathbb{R}}^{1024\times 1}$
      \State binarizing an image of NIST to obtain $\boldsymbol{z}$
      \State extracting the central part of $\boldsymbol{z}$ to obtain $\boldsymbol{z}_c$ with dimension $64 \times 64$
      \State downsampling $\boldsymbol{z}_c$ to obtain $\boldsymbol{z}_{cd}$ with dimension $32 \times 32$
      \State extracting the edge of $\boldsymbol{z}_{cd}$ to obtain the mask image $\boldsymbol{m}_{cd}$
      \State decomposing $\boldsymbol{m}_{cd}$ into four parts $\boldsymbol{m}_{\text{outside}}$, $\boldsymbol{m}_{\text{outside-boundary}}$, $\boldsymbol{m}_{\text{inside-boundary}}$, and $\boldsymbol{m}_{\text{inside}}$.
     \State sorting $\boldsymbol{x}$ in the descending order to derive $\boldsymbol{\hat{x}}$
     \State decomposing $\boldsymbol{\hat{x}}$ into four parts, i.e., $\boldsymbol{\hat{x}} = \{ \boldsymbol{\hat{x}}_1, \boldsymbol{\hat{x}}_2, \boldsymbol{\hat{x}}_3, \boldsymbol{\hat{x}}_4\}$
     \State Placing each pixel of $\{ \boldsymbol{\hat{x}}_1, \boldsymbol{\hat{x}}_2, \boldsymbol{\hat{x}}_3, \boldsymbol{\hat{x}}_4\}$ into a random position within the corresponding masks to generate a synthetic image.
    \Until{(20,000 synthetic images are generated)}     
    
    \endgroup
    \Ensure  
      The synthetic dataset    
  \end{algorithmic}  
\end{algorithm}

\section{Experiments}
\label{sec:pagestyle}

In this section, we demonstrate that the synthetic dataset can be easily used to DNNs.
First, we design a simple but comprehensive Convolutional Neural Network (abbr. CNN1) for classifying the synthetic dataset.
CNN1 has five hidden layers: two convolutional layers, two ReLU operator, and two max pooling layers.
Table \ref{cnns_synthetic} summarizes the architecture of CNN1.

We take 30 training epochs to train CNN1 for classifying the synthetic dataset, and the learning rate is 0.008.
Fig. \ref{fig_cnn_synthetic} shows the performance of CNN1 on the synthetic dataset.
We can see that CNN1 achieves zero training error after 20 training epochs, the testing error is also very small.
Overall, we can conclude that the synthetic dataset can be applied to DNNs

\begin{table}[!b]
\caption{The architectures of CNN1 for experiments}
\label{cnns_synthetic}
\vskip 0.15in
\begin{center}
\begin{small}
\begin{threeparttable}
\begin{tabular}{cccc}
\toprule
R.V. & Layer & Description & CNN1 \\
\midrule
$\boldsymbol{X}$&$\boldsymbol{x}$					& Input 				& $32 \times 32 \times 1$ \\
\hline
$\boldsymbol{F_1}$ &$\boldsymbol{f_1}$				& Conv ($3 \times 3$)	& $30 \times 30 \times {20}$ \\
\hline 
\multirow{2}*{$\boldsymbol{F_2}$}&$\boldsymbol{f_2}$    	& Maxpool + ReLU		& $15 \times 15 \times {20}$ \\
&$\boldsymbol{f_3}$    							& Conv ($5 \times 5$)	& $11 \times 11 \times {60}$ \\
\hline 
\multirow{3}*{$\boldsymbol{F_Y}$}&$\boldsymbol{f_4}$    	& Maxpool + ReLU		& $5 \times 5 \times {60}$  \\
&$\boldsymbol{f_5}$      							& Fully connected		& $1 \times 1 \times 10$  \\
&$\boldsymbol{f_Y}$   							& Output(softmax)		& $1 \times 1 \times 10$  \\
\bottomrule
\end{tabular}
\begin{tablenotes}
            \item R.V. is the random variable of the hidden layer(s).
   \end{tablenotes}
\end{threeparttable}
\end{small}
\end{center}
\vskip -0.1in
\end{table}

\section{Conclusion}
\label{sec:ref}

In this work, we propose a novel method for generating a synthetic dataset.
In contrast to the commonly used benchmark datasets with unknown distribution, the synthetic dataset has a explicit distribution, i.e., Gaussian distribution.
In particular, it has the same characteristics of the benchmark dataset MNIST.
As a result, we can easily apply Deep Neural Networks (DNNs) on the synthetic dataset.


\bibliographystyle{IEEEbib}
\bibliography{mlsp_2019_ethan}

\begin{thebibliography}{10}

\bibitem{CNN-Nature}
Yann LeCun, Yoshua Bengio, and Geoffrey Hinton,
\newblock ``Deep learning,''
\newblock {\em Nature}, pp. 436--444, 2015.

\bibitem{speech-recognition}
G.~E. Hinton, D.~Li, Y.~Dong, E.~George, A.~Mohamed, N.~Jaitly, A.~Senior,
  V.~Vanhoucke, P.~Nguyen, T.~Sainath, and B.~Kingsbury,
\newblock ``Deep neural networks for acoustic modeling in speech recognition,''
\newblock {\em IEEE Signal Processing Magazine}, vol. 29, pp. 82--97, 2012.

\bibitem{CNN}
Krizhevsky, Sutskever A., and G.~E. Hinton,
\newblock ``Imagenet classification with deep convolutional neural networks,''
\newblock in {\em NeurIPS}, 2012, vol.~25, pp. 1090--1098.

\bibitem{DNN_blackbox}
Guillaume Alain and Yoshua Bengio,
\newblock ``Understanding intermediate layers using linear classifier probes,''
\newblock {\em arXiv preprint arXiv:1610.01644}, 2016.

\bibitem{posterior1}
Herbert Gish,
\newblock ``A probabilistic approach to the understanding and training of
  neural network classifiers,''
\newblock in {\em IEEE ICASSP}, 1990, pp. 1361--1364.

\bibitem{pearl}
Judea Pearl,
\newblock ``Theoretical impediments to machine learning with seven sparks from
  the causal revolution,''
\newblock {\em arXiv preprint arXiv:1801.04016}, 2018.

\bibitem{cnn-posterior}
M.D. Richard and R.P. Lippmann,
\newblock ``Neural network classifiers estimate bayesian a posteriori
  probabilities,''
\newblock {\em Neural Computation}, pp. 461--483, 1991.

\bibitem{posterior4}
G.~Zhang,
\newblock ``Neural networks for classification: a survey,''
\newblock {\em IEEE Transactions on Systems, Man, and Cybernetics}, vol. 30,
  pp. 451--462, 2000.

\bibitem{representation-dl}
Yoshua Bengio, Aaron Courville, and Pascal Vincent,
\newblock ``Representation learning: A review and new perspectives,''
\newblock {\em IEEE Transactions on Pattern Analysis and Machine Intelligence},
  vol. 35, no. 8, pp. 1798--1828, 2013.

\bibitem{Goodfellow-et-al-2016}
Ian Goodfellow, Yoshua Bengio, and Aaron Courville,
\newblock {\em Deep Learning},
\newblock MIT Press, 2016.

\bibitem{DRMM}
Ankit Patel, Minh Nguyen, and Richard Baraniuk,
\newblock ``A probabilistic framework for deep learning,''
\newblock in {\em NeurIPS}, 2016.

\bibitem{Soatto1}
Alessandro Achille and Stefano Soatto,
\newblock ``Emergence of invariance and disentanglement in deep
  representations,''
\newblock {\em arXiv preprint arXiv:1706.01350}, 2017.

\bibitem{DNN-Bottleneck}
Noga~Zaslavsky Naftali~Tishby,
\newblock ``Deep learning and the information bottleneck principle,''
\newblock {\em arXiv preprint arXiv:1503.02406}, 2015.

\bibitem{IP-argue}
Andrew Saxe, Yamini Bansal, Joel Dapello, Madhu Advani, Artemy Kolchinsky,
  Brendan Tracey, and David Cox,
\newblock ``On the information bottleneck theory of deep learning,''
\newblock in {\em ICLR}, 2018.

\bibitem{DNN-information}
Ravid Shwartz-Ziv and Naftali Tishby,
\newblock ``Opening the black box of deep neural networks via information,''
\newblock {\em arXiv preprint arXiv:1703.00810}, 2017.

\bibitem{LeCun}
Y.~LeCun, L.~Bottou, Y.~Bengio, and P.~Haffner,
\newblock ``Gradient-based learning applied to document recognition,''
\newblock {\em Proceedings of the IEEE}, vol. 11, no. 86, pp. 2278--2324,
  November 1998.

\bibitem{xinjie2}
Xinjie Lan and Kenneth~E. Barner,
\newblock ``From mrfs to cnns: A novel image restoration method,''
\newblock in {\em 52nd Annual Conference on Information Sciences and Systems
  (CISS)}, 2018, pp. 1--5.

\bibitem{CRF_RNN}
Shuai Zheng, Sadeep Jayasumana, Bernardino Romera-Paredes, Vibhav Vineet,
  Zhizhong Su, Dalong Du, Chang Huang, and Philip Torr,
\newblock ``Conditional random fields as recurrent neural networks,''
\newblock in {\em International Conference on Computer Vision (ICCV)}, 2015,
  pp. 1529--1537.

\bibitem{statistical_img}
E.~P. Simoncelli,
\newblock ``Statistical models for images: Compression, restoration and
  synthesis,''
\newblock in {\em Proc 31st Asilomar Conf on Signals, Systems and Computers},
  November 1997, pp. 673--678.

\bibitem{GSM-stat}
Martin.~J. Wainwright and Eero.~P. Simoncelli,
\newblock ``Scale mixtures of gaussians and the statistics of natural images,''
\newblock in {\em NeurIPS}, 2000, pp. 855--861.

\bibitem{Canny}
Lijun Ding and Ardeshir Goshtasby,
\newblock ``On the canny edge detector,''
\newblock {\em Pattern Recognition}, vol. 34, pp. 721--725, 2001.

\end{thebibliography}

\end{document}